\documentclass{llncs}
\usepackage{makeidx}  
\usepackage{graphicx}
\usepackage{caption}
\usepackage{amsmath,amssymb}
\usepackage{wrapfig}
\usepackage{color}
\usepackage{subfig}
\usepackage{footmisc}

\newcommand{\jointfirstauthor}{*}

\begin{document}
\mainmatter              
\title{Nonlinear Spectral Image Fusion}
\titlerunning{Spectral Image Fusion}  %
\author{Martin Benning\inst{1}\textsuperscript{,\jointfirstauthor} \and Michael M\"{o}ller\inst{2}\textsuperscript{,\jointfirstauthor} \and Raz Z. Nossek\inst{3}\textsuperscript{,\jointfirstauthor} \and Martin Burger\inst{4} \and\\ Daniel Cremers\inst{5} \and Guy Gilboa\inst{3} \and Carola-Bibiane Sch\"{o}nlieb\inst{1}}
\institute{University of Cambridge, Wilberforce Road, Cambridge, CB3 0WA, UK\\ \email{\{mb941, cbs31\}@cam.ac.uk} \and Universit\"{a}t Siegen, H\"{o}lderlinstra\ss e 3, 57076 Siegen, Germany\\ \email{michael.moeller@uni-siegen.de} \and Technion IIT, Technion City, Haifa 32000, Israel\\ \email{\{nossekr@campus, guy.gilboa@ee\}.technion.ac.il} \and Westf\"{a}lische Wilhelms-Universit\"{a}t, Einsteinstrasse 62, 48149 M\"{u}nster, Germany\\ \email{martin.burger@wwu.de} \and Technische Universit\"{a}t M\"{u}nchen, Boltzmannstrasse 3, 85748 Garching, Germany\\ \email{cremers@tum.de}}

\maketitle  

\begin{abstract}
In this paper we demonstrate that the framework of nonlinear spectral decompositions based on total variation (TV) regularization is very well suited for image fusion as well as more general image manipulation tasks. The well-localized and edge-preserving spectral TV decomposition allows to select frequencies of a certain image to transfer particular features, such as wrinkles in a face, from one image to another. We illustrate the effectiveness of the proposed approach in several numerical experiments, including a comparison to the competing techniques of Poisson image editing, linear osmosis, wavelet fusion and Laplacian pyramid fusion. We conclude that the proposed spectral TV image decomposition framework is a valuable tool for semi- and fully-automatic image editing and fusion.
\keywords{Nonlinear spectral decomposition, total variation regularization, image fusion, image composition, multiscale methods}
\end{abstract}\footnotetext[1]{These authors contributed equally to this work.}

\section{Introduction}
\label{sec:intro}

\begin{figure}[!ht]
  \centering
    \includegraphics[width=1\textwidth]{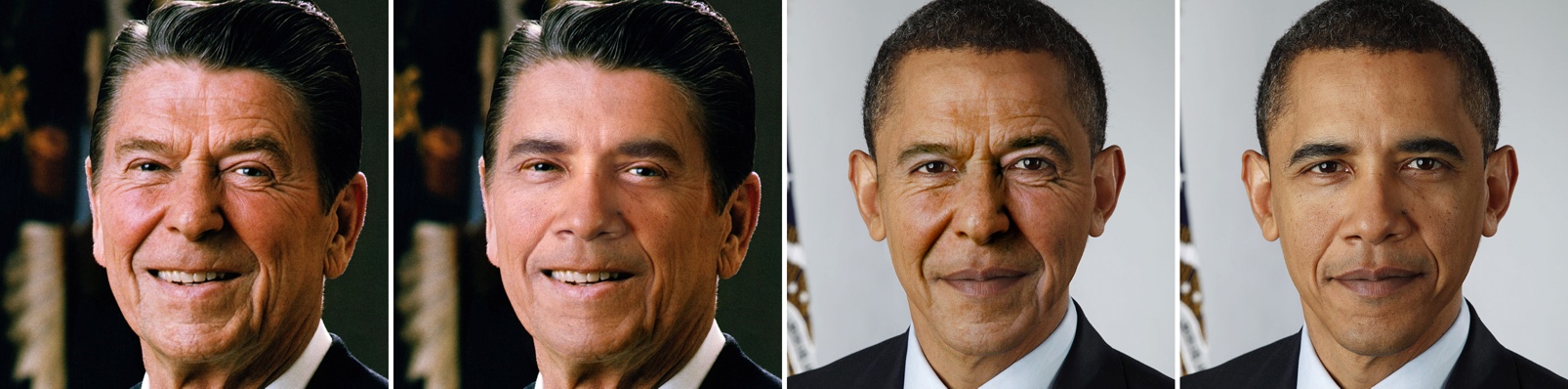}
     \caption{Example of a facial image fusion result obtained by the proposed framework. The left image of Reagan and right image of Obama were used as an input, while the two middle images are synthesized using nonlinear spectral TV filters.}
     \label{fig:teaser}
\end{figure}

Since the rise of digital photography people have been fascinated by the possibilities of manipulating digital images. In this paper we present an image manipulation and fusion framework based on the recently proposed technique of nonlinear spectral decompositions \cite{Gilboa_SSVM_2013_SpecTV,Gilboa_spectv_SIAM_2014,spec_one_homog15} using TV regularization. By defining spectral filters that extract features corresponding to particular frequencies, we can for instance transfer wrinkles from one face to another and create visually convincing fusion results as shown in Figure \ref{fig:teaser}.

Classical multiscale methods such as Fourier analysis, sine or cosine transformations, or wavelet decompositions represent an input image as a linear superposition of a given set of basis elements. In many cases, these basis elements are given by the eigenfunctions of a suitable linear operator. For instance, the classical Fourier representation of a function as a superposition of sine and cosine functions arises from the eigenfunctions of the Laplace operator, i.e. from functions $v_\lambda$ with $\|v_\lambda\|=1$ and $\Delta v_\lambda = \lambda v_\lambda$, with periodic boundary conditions. Interestingly, the condition for $v_\lambda$ being an eigenfunction can be written in terms of the regularization functional $J(v) = \frac{1}{2}\|\nabla v\|_2^2$ as
\begin{align}
\label{eq:eigen}
\lambda v_\lambda \in \partial J(v_\lambda),
\end{align}
where $ \partial J(v) = \{p \in \mathcal{X}^*  ~|~ J(u) - J(v) - \langle p, u - v \rangle \geq 0 \}$ denotes the subdifferential of the functional $J: \mathcal{X} \rightarrow \mathbb{R}$, with $\mathcal{X}$ being a suitable function space, typically a Banach space. Since inclusion \eqref{eq:eigen} makes sense for arbitrary convex regularization functions (e.g. for TV regularization), it provides a natural definition for generalizing the concept of eigenfunctions, cf. \cite{Benning_Burger_2013}. 

The idea of nonlinear spectral decompositions \cite{Gilboa_SSVM_2013_SpecTV,Gilboa_spectv_SIAM_2014,spec_one_homog15} (which we will recall in more detail in Section \ref{sec:spectralDecompo}) is built upon the idea that an eigenfunction in the spatial domain, i.e. an element meeting \eqref{eq:eigen}, should be represented as a single peak in the spectral domain. Decompositions with respect to TV regularization have been shown to provide a highly image-adaptive way to represent different features and scales, see \cite{Gilboa_SSVM_2013_SpecTV}. 

In this paper we will demonstrate that the nonlinear spectral image decomposition framework is very well suited for several challenging image fusion tasks. Our contributions include
\begin{itemize}
\item Proposing a nonlinear spectral image editing and fusion framework.
\item Providing a robust pipeline for the automatic fusion of human faces, including face and landmark detection, registration, and segmentation.
\item Illustrating state-of-the-art results evaluated against Laplacian pyramid fusion, wavelet fusion, Poisson image editing, and linear osmosis.
\item Demonstrating the flexibility of the proposed framework beyond the fusion of facial images by considering applications such as object insertion and image style manipulation.
\end{itemize}

\textbf{Copyright remark: }{All photographs used in this paper were taken from the Wikipedia Commons page, \url{https://commons.wikimedia.org/}, or from the free images site \url{https://commons.pixabay.com/}. The photo of Barack Obama was made by Pete Souza - Creative Commons Attribution 3.0 Unported license, see \url{https://creativecommons.org/licenses/by/3.0/deed.en}}

\section{Image Fusion}
\label{sec:relatedWork}
The most common image fusion techniques use a multiscale approach such as wavelet decompositions \cite{zeeuw1998wavelets} or a Laplacian pyramid \cite{burt1983multiresolution} to decompose two or more images, combine the decompositions differently on different scales, and reconstruct an image from the fused multiscale decomposition. Applications of the aforementioned fusion techniques include generating an all-in-focus image from a stack of differently focused images (e.g. \cite{Li2008}), multi- and hyperspectral imagery (cf. \cite{Amolins07}), or facial texture transfers \cite{thies2015realtime}.

It was shown in \cite{spec_one_homog15,Gilboa16} that the nonlinear spectral decomposition framework actually reduces to the usual wavelet decomposition when the TV regularization is replaced by $J(u) = \|Wu\|_1$, where $W$ denotes the linear operator conducting the (orthogonal) wavelet transform. We, however, are going to demonstrate that the image-adaptive nonlinear decomposition approach with TV regularization is significantly better suited for image manipulation and fusion tasks.

Several other sophisticated nonlinear image multiscale decompositions have been proposed including techniques based on bilateral filtering (e.g. \cite{Fattal07}), weighted least-squares \cite{Farbman08}, local histograms \cite{Kass10}, local extrema \cite{Subr09}, or Gaussian structure--texture decomposition \cite{Su}. Applications of the aforementioned methods include image equalization and abstraction, detail enhance or removal, and tone mapping/manipulation. While \cite{Su} briefly discusses applications in texture transfer, the potential of a complete image fusion by combining different frequencies of different images has not been exploited sufficiently yet.

For various image editing tasks related to inserting objects from one image $g$ into another image $f$, the seminal work of Perez, Gangnet and Blake on Poisson image editing \cite{poissonImageEditing}, provides a valuable tool. The authors proposed to minimize  $E(u) = \|\nabla u - \nabla g\|^2$ subject to $u$ coinciding with $f$ outside of the region the object is to be inserted into. 
%

Recent improvements of the latter have been made with osmosis image fusion, cf. \cite{linearOsmosis,osmosisFilters}. Linear osmosis filtering for image fusion is achieved by solving a drift-diffusion PDE; here the drift vector field is constructed by combining the two vector fields $\nabla \ln(g)$ and $\nabla \ln(f)$; parts of $\nabla \ln(g)$ are inserted into $\nabla \ln(f)$, and averaged across the boundary. The initial value of the PDE is set to $f$, or the mean of $f$. A detailed description of the procedure is given in \cite[Section 4.3]{linearOsmosis}. For a general overview of image fusion techniques in different areas of application we also refer the reader to \cite{Stathaki08}.

\section{Nonlinear Spectral Fusion}
\label{sec:fusion}
The starting point and motivation for extending linear multiscale methods such as Fourier or wavelet decompositions into a nonlinear setting are basis elements, which often originate as eigenfunctions of a particular linear operator. As shown in Section \ref{sec:intro}, Fourier analysis can be recovered by decomposing a signal into a superposition of elements $v_\lambda$ meeting the inclusion \eqref{eq:eigen}.

As mentioned in the introduction, the disadvantage of conventional decomposition techniques is the lack of adaptivity of the basis functions. In the following, we recall the definition of more general, nonlinear spectral transformations that allow to create more adaptive decompositions of images.

\subsection{Nonlinear Spectral Decomposition}
\label{sec:spectralDecompo}
The idea of nonlinear spectral decompositions of \cite{Gilboa_SSVM_2013_SpecTV,Gilboa_spectv_SIAM_2014,spec_one_homog15} is to consider \eqref{eq:eigen} for one-homogeneous functionals $J$ (such as TV) instead of quadratic ones, which give rise to classical multiscale image representations. Since eigenvectors of one-homogeneous functionals are difficult to compute numerically (cf. \cite{Benning_Burger_2013}), the property one aims to preserve is that input data given in terms of an eigenfunction is decomposed into a single peak when being transformed into its corresponding (nonlinear) frequency representation.

Let us consider an eigenfunction $f = v_\lambda$, $\|v_\lambda\|_2=1$, obeying \eqref{eq:eigen}, and consider the behavior of the gradient flow
\begin{align}
\label{eq:gradientFlow}
\partial_t u_{GF}(t) = -p_{GF}(t), \qquad p_{GF}(t) \in \partial J(u_{GF}(t)), \qquad u_{GF}(0) = f,
\end{align}
for a one-homogeneous functional $J$. It follows almost directly from \cite[Theorem 5]{Benning_Burger_2013} that the solution to this problem is given by
\begin{align}
\label{eq:eigenfunctionDecomposition}
 u_{GF}(t) = \left \{
\begin{array}{ll}
(1-t\lambda) f & \text{ if } t\lambda\leq 1, \\
0 & \text{ else. }
\end{array}
\right.
\end{align}
Since $u_{GF}(t)$ behaves piecewise linear in $t$, one can consider the second derivative to obtain a $\delta$-peak. One defines
\begin{align}
\label{eq:TvTrans}
\phi_{GF}(t) = t \partial_{tt} u_{GF}(t)
\end{align}
to be the spectral decomposition of the input data $f$, even in the case where $f$ is not an eigenfunction of $J$. The additional normalization factor $t$ admits to the reconstruction formula
\begin{align}
\label{eq:ReconTvTrans}
f = \int_0^\infty \phi_{GF}(t)  ~dt + \overline{f},
\end{align}
with $\overline{f} := \min_{\tilde{f} \in \text{kernel}(J)} \| \tilde{f} - f \|_2$, for arbitrary $f$. We refer the reader to \cite{Gilboa_spectv_SIAM_2014} for more details on the general idea, and to \cite{BEGM_1hom_SIAM_submitted} for a mathematical analysis of the above approach.

As we can see in \eqref{eq:eigenfunctionDecomposition}, peaks of eigenfunctions in $\phi_{GF}$ appear at $t = \frac{1}{\lambda}$, i.e. earlier the bigger $\lambda$ is. Therefore, one can interpret $\phi_{GF}$ as a wavelength decomposition, and motivate wavelength based filtering approaches of the form
\begin{align}
\label{eq:FilterTvTrans}
\hat{u} = \int_0^\infty H(t) \ \phi_{GF}(t)  ~dt +  \overline{H} \; \overline{f},
\end{align}
where the filter function $H$ (along with the weight $\overline{H}$) can enhance or suppress selected parts of the spectrum.

As discussed in \cite{spec_one_homog15}, there exists an alternative formulation to the gradient flow representation defined in \eqref{eq:gradientFlow}. One can also consider the inverse scale space flow (see \cite{iss,burger2007inverse})
\begin{align}
\label{eq:invscalspaceflow}
\partial_t p_{IS}(t) = f - u_{IS}(t), \quad p_{IS}(t) \in \partial J(u_{IS}(t)), \quad p_{IS}(0) = 0.
\end{align}
For certain regularizations $J$, the two approaches are provably equivalent (cf. \cite{BEGM_1hom_SIAM_submitted}); hence, we use the approaches interchangeably based on the numerical convenience, as we also empirically observe very little difference between the numerical realisations of \eqref{eq:gradientFlow} and \eqref{eq:invscalspaceflow}.

Note that we use the total variation as the regularizer $J$ throughout the remainder of this paper; however, other choices for $J$ are possible (see \cite{spec_one_homog15}).

\subsection{Numerical Implementation}
\subsubsection{Spectral Decomposition}
For the numerical implementation of our spectral image fusion we use both the gradient flow as well as the inverse scale scale flow formulation. The former is implemented in the exact same way as described in \cite{Gilboa_spectv_SIAM_2014}. Formulation \eqref{eq:invscalspaceflow} is discretized via Bregman iterations (cf. \cite{Bregman_obgxy}). More precisely, we compute
\begin{align}
\label{eq:variationalBregman}
u^{k+1} &= \arg \min_u \frac{\tau^{k+1}}{2}\|u-f\|_2^2 + (TV(u) - \langle p^k, u \rangle) ,\\
\label{eq:pUpdate}
p^{k+1} &= p^k + \tau^{k+1}(f - u^{k+1}),
\end{align}
starting with $p^0=0$. We then define
\begin{align}
\psi^k = \left\{ \begin{array}{ll}u^1 & \text{ if } k=1, \\  u^{k}-u^{k-1} & \text{ else,}
\end{array} \right.
\end{align}
to be the frequency decomposition of the input data $f$.

\begin{figure}[!ht]
  \centering
    \includegraphics[width=1\textwidth]{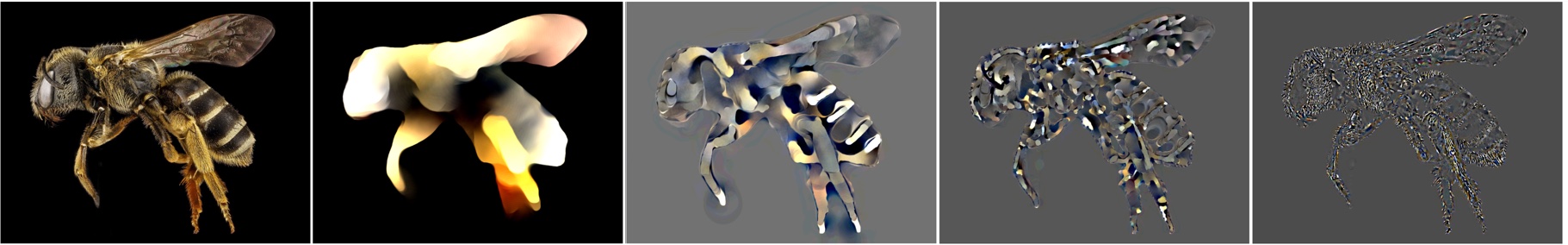}
     \caption{Example of a nonlinear frequency decomposition. The left image is the input image to be decomposed, the following images illustrate selected spectral components with increasing associated frequencies. This type of decomposition is the main tool for our proposed image fusion framework.}
     \label{fig:decomposition}
\end{figure}

From the optimality condition of equation \eqref{eq:variationalBregman} we conclude that $p^k \in \partial TV(u^k)$ for all $k$. Furthermore, note that equation \eqref{eq:pUpdate} can be rewritten as
\begin{align}
\label{eq:ReformP_Update}
\frac{p^{k+1}-p^k}{\tau^{k+1}} = f - u^{k+1}
\end{align}
and can therefore be interpreted as the discretization of the inverse scale space flow. In our numerical implementation we use the adaptive step size $\frac{1}{\tau^k} = 30 \cdot 0.6^{k-1}$ to better resolve significant changes of the flow. With this adaptation, we found 15 iterations to be sufficient to approximately converge to $u^{15}=f$ and to still obtain a sufficiently detailed frequency decomposition. Figure \ref{fig:decomposition} illustrates a generalized frequency representation using the above method on an input image of a bee.

To solve the minimization problem of equation \eqref{eq:variationalBregman} numerically we use the primal-dual hybrid gradient method with diagonal preconditioning \cite{Pock-Chambolle-iccv11} and the adaptive step size rule from \cite{Goldstein-Esser-13}.

\subsubsection{Image Fusion}
The general idea of the spectral image fusion is to apply the nonlinear spectral image decomposition to two images or regions therein, combine the coefficients at different scales, and reconstruct an image from the fused coefficients. 

Let $v$ be a registration function that aligns a part of the second image with the location in the first image where the object is to be inserted into. Given the corresponding spectral decompositions $\phi^1$ and $\phi^2$, we compute the fused image $u_{\text{fused}}$ via
\begin{align}
\label{eq:fusion}
u_{\text{fused}}(x) = \int_0^\infty H^1(x,t) \phi^1(x,t) +  H^2(x+v(x),t) \phi^2(x+v(x),t) ~dt,
\end{align}
where the two filter functions $H^1$ and $H^2$ determine the amount of spectral information to be included in the fused image. Finally, we add a weighted linear combination of the constant parts $\overline{f^1}$ and $\overline{f^2}$ of the two input images $f^1$ and $f^2$ to $u_{\text{fused}}$. Note that -- opposed to the original spectral representation framework from \cite{Gilboa_SSVM_2013_SpecTV,Gilboa_spectv_SIAM_2014,spec_one_homog15} -- we are considering $x$-dependent, i.e. \textit{spatially varying filters}, to adapt the filters in different regions of the images.

\begin{figure}[!ht]
  \centering
    \includegraphics[width=0.9\textwidth]{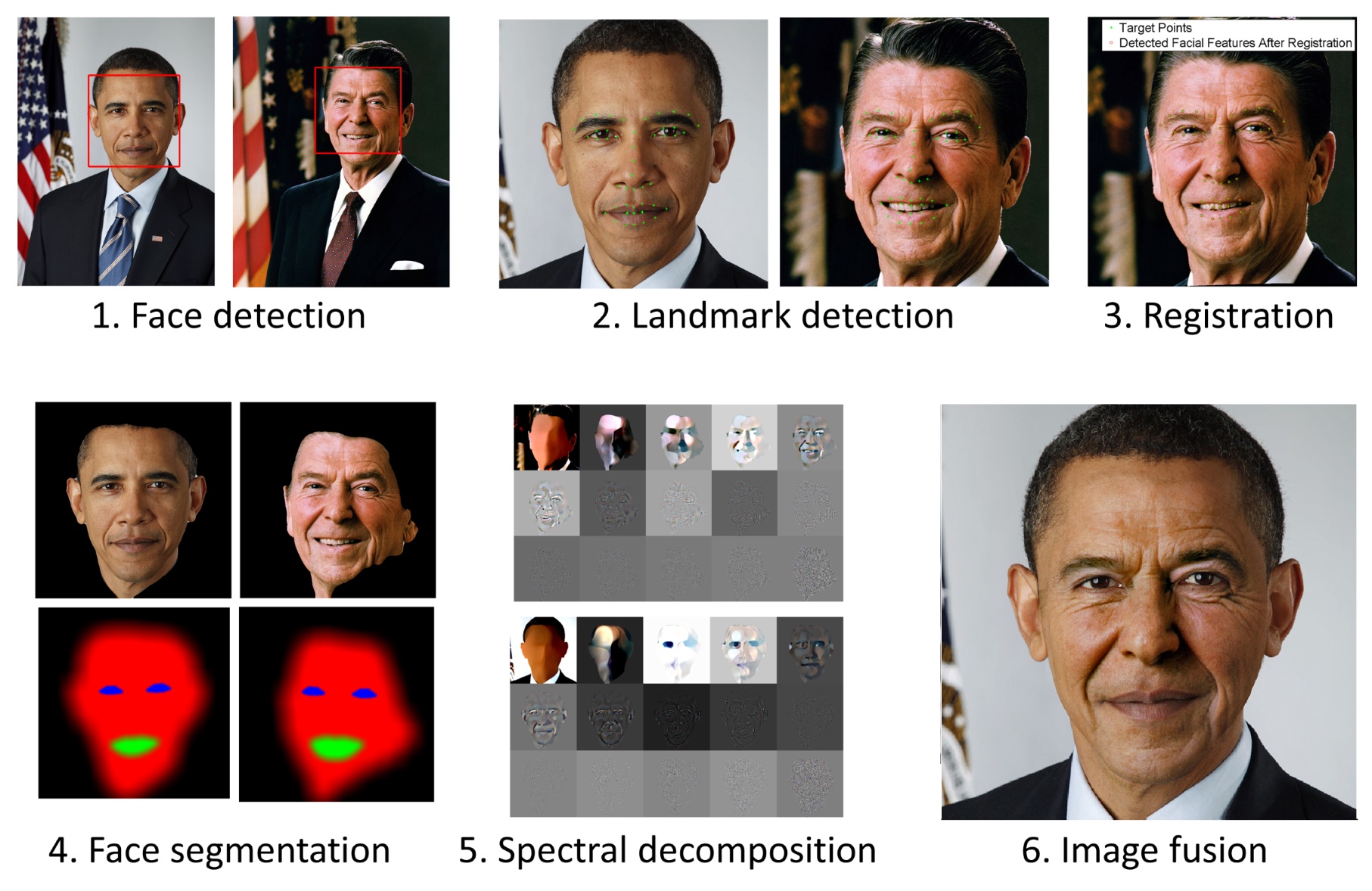}
     \caption{Illustration of the pipeline for facial image fusion using nonlinear spectral decompositions.}
     \label{fig:concept}
\end{figure}

\section{Results}
\label{sec:results}
\subsection{Automatic Image Fusion of Human Faces}
\label{sec:facesresults}
To illustrate the concept of using nonlinear spectral decompositions for image editing, we consider the problem of fusing two images of human faces. The latter has attracted quite some attention in the literature before, see e.g. \cite{blanz1999morphable,thies2015realtime}. Note that in contrast to \cite{blanz1999morphable,thies2015realtime} our fusion process does not depend on a 3d model of a face (which naturally means our framework does not handle changes of perspective). 

For the presented image fusion, we have developed a fully automatic image fusion pipeline illustrated in Figure \ref{fig:concept}. It consists of face detection using the Viola-Jones algorithm \cite{ViolaJones}, facial landmark detection using \cite{Asthana14}, determining the non-rigid registration field $v$ that has a minimal Dirichlet energy $\|\nabla v\|_2$ among all

possible maps that register the detected landmarks, a face segmentation using the approach in \cite{Nieuwenhuis12} with additional information from the landmarks to distinguish between the face, mouth, and eye region, and finally the decomposition and fusion steps described in Section \ref{sec:fusion}, where we restrict the decomposition to the regions of interest to be fused. Upon acceptance of this paper we will make the source code available in order to provide more details of the implementation.

\begin{figure}[!ht]
  \centering
    \includegraphics[width=0.9\textwidth]{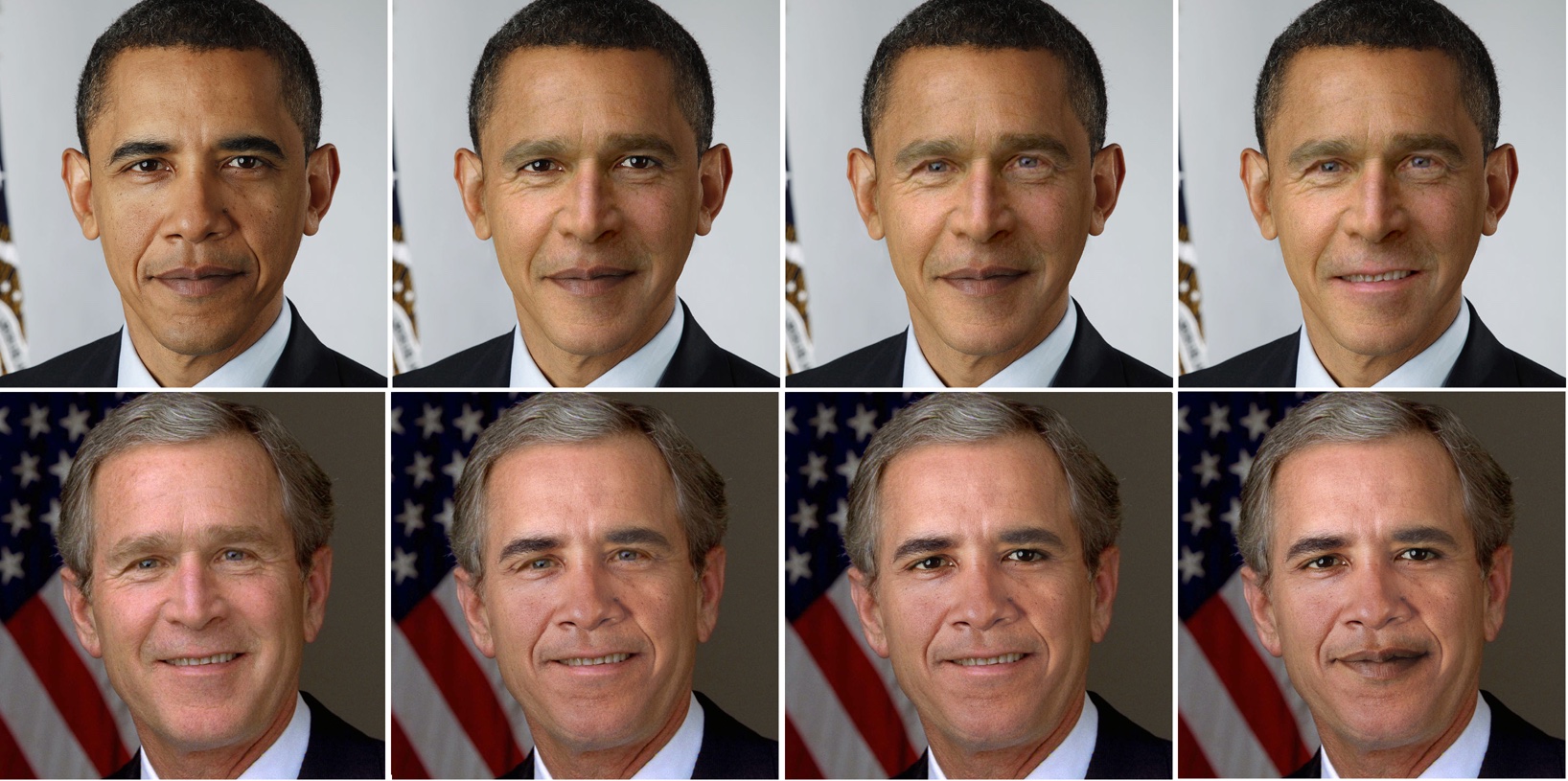}
\caption{The sub-segmentation of each image into a face, a mouth and an eye region allows to define spatially varying filters. The above images illustrate effects of incorporating eyes or the mouth from one or the other image.}
     \label{fig:eyeMouthIllustration}
\end{figure}

The segmentation into the subregions allows us to define spatially varying spectral filters that treat the eye, mouth, and remaining facial regions differently, where fuzzy segmentation masks are used to blend the spectral filters from one region into the next to create smooth and visually pleasing transitions. Effects one can achieve by varying the spectral filters in the eye and mouth regions are illustrated in Figure \ref{fig:eyeMouthIllustration}.

\begin{figure}[!ht]
  \centering
    \subfloat[Face filter for first image]{
    \includegraphics[width=0.47\textwidth]{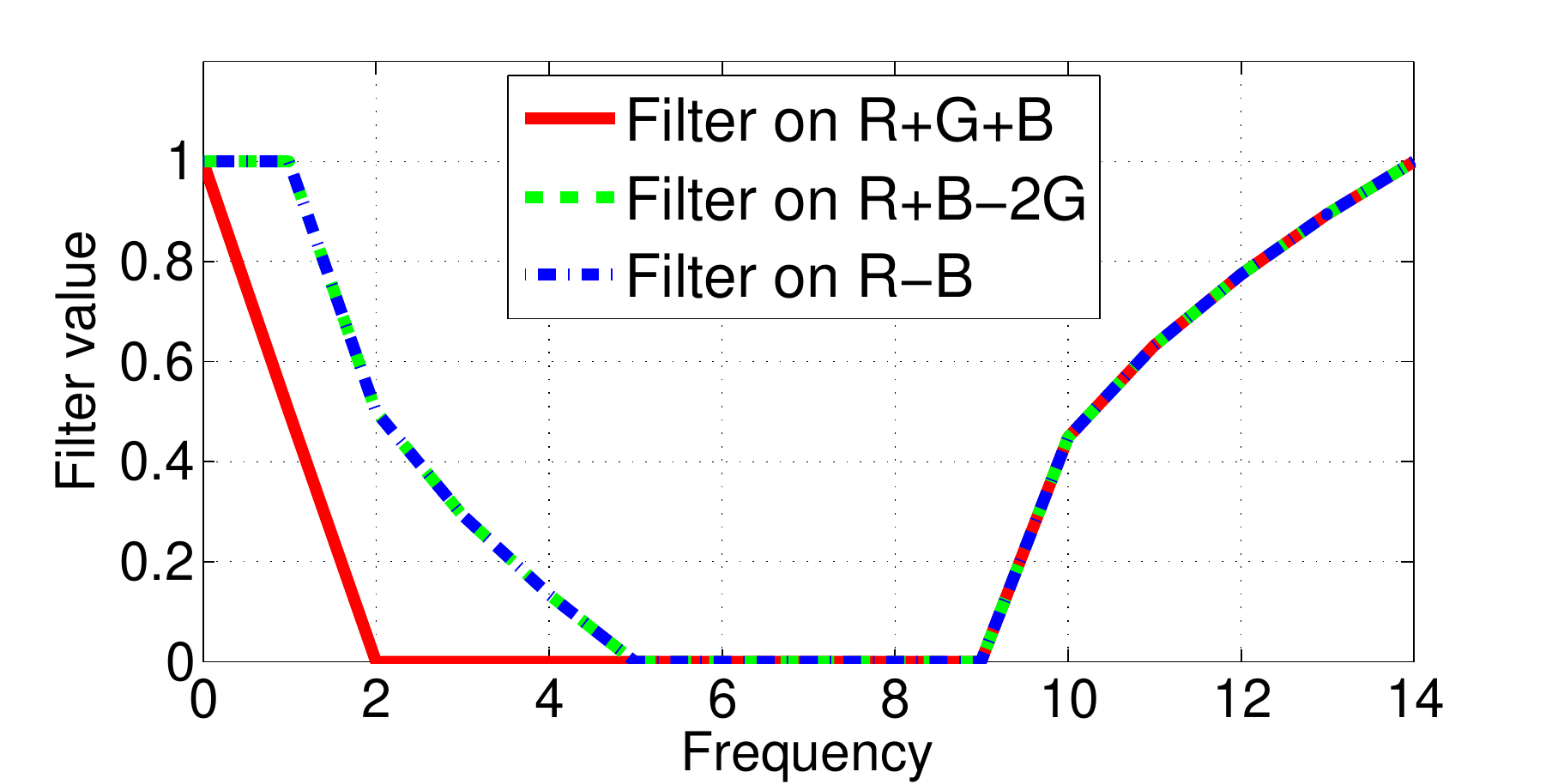}
    }
   \subfloat[Face filter for second image]{
    \includegraphics[width=0.47\textwidth]{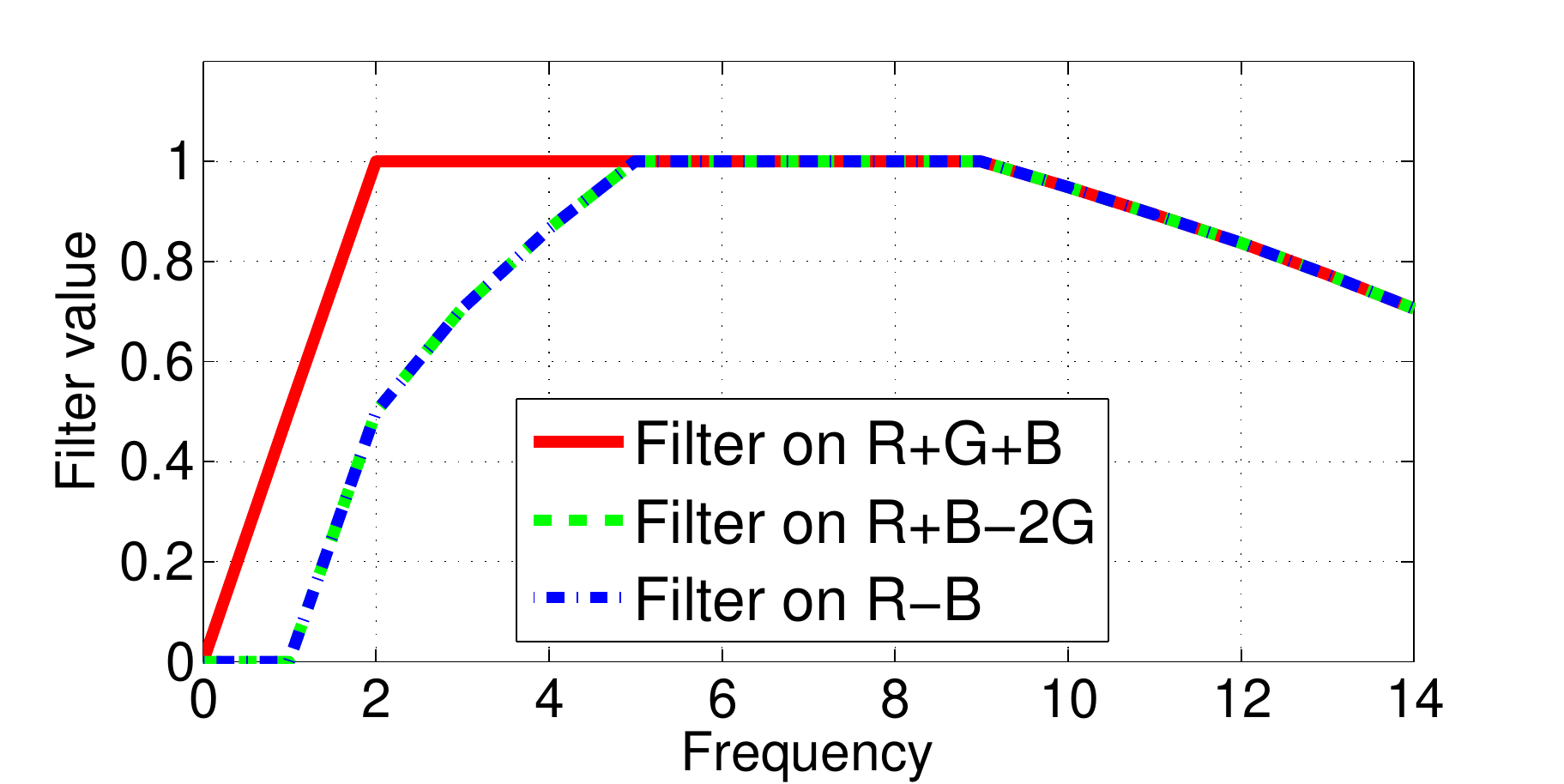}
    }
     \caption{Illustration of the fusion filters to generate the images in Figure \ref{fig:teaser}.}
     \label{fig:ObambaRegeanFilters}
\end{figure}

Figure \ref{fig:ObambaRegeanFilters} shows the filters we used to fuse the faces of the presidents Obama and Reagan for the introductory example in Figure \ref{fig:teaser}. As illustrated, the spectral filters may also differ for each of the color channels and can therefore also be applied to images decomposed into luminance and chrominance channels. In our examples we used the $(L, C^{G/M}, C^{R/B})$ color transform which has shown a promising performance e.g. for image demosaicking in \cite{condat12}. As we can see in Figure \ref{fig:ObambaRegeanFilters}, one might want to keep more chrominance values of the target image to retain similar color impressions. Furthermore, the filter responses do not have to sum to one. In the high frequencies we keep a good amount of both images, which -- in our experience -- leads to sharper and more appealing results with skin-textures from both images.

To illustrate the robustness of the proposed framework, we ran the fully automatic image fusion pipeline on an image set of US presidents gathered from the Wikipedia Commons page. The results are shown in the supplementary material accompanying this manuscript. The proposed nonlinear image fusion approach is robust enough to work with a great variety of different images and types of photos. The supplementary material contains further examples of fusing people with statues, and fusing a bill with a painting.

\begin{figure}[!h]
\includegraphics[width=0.32\textwidth]{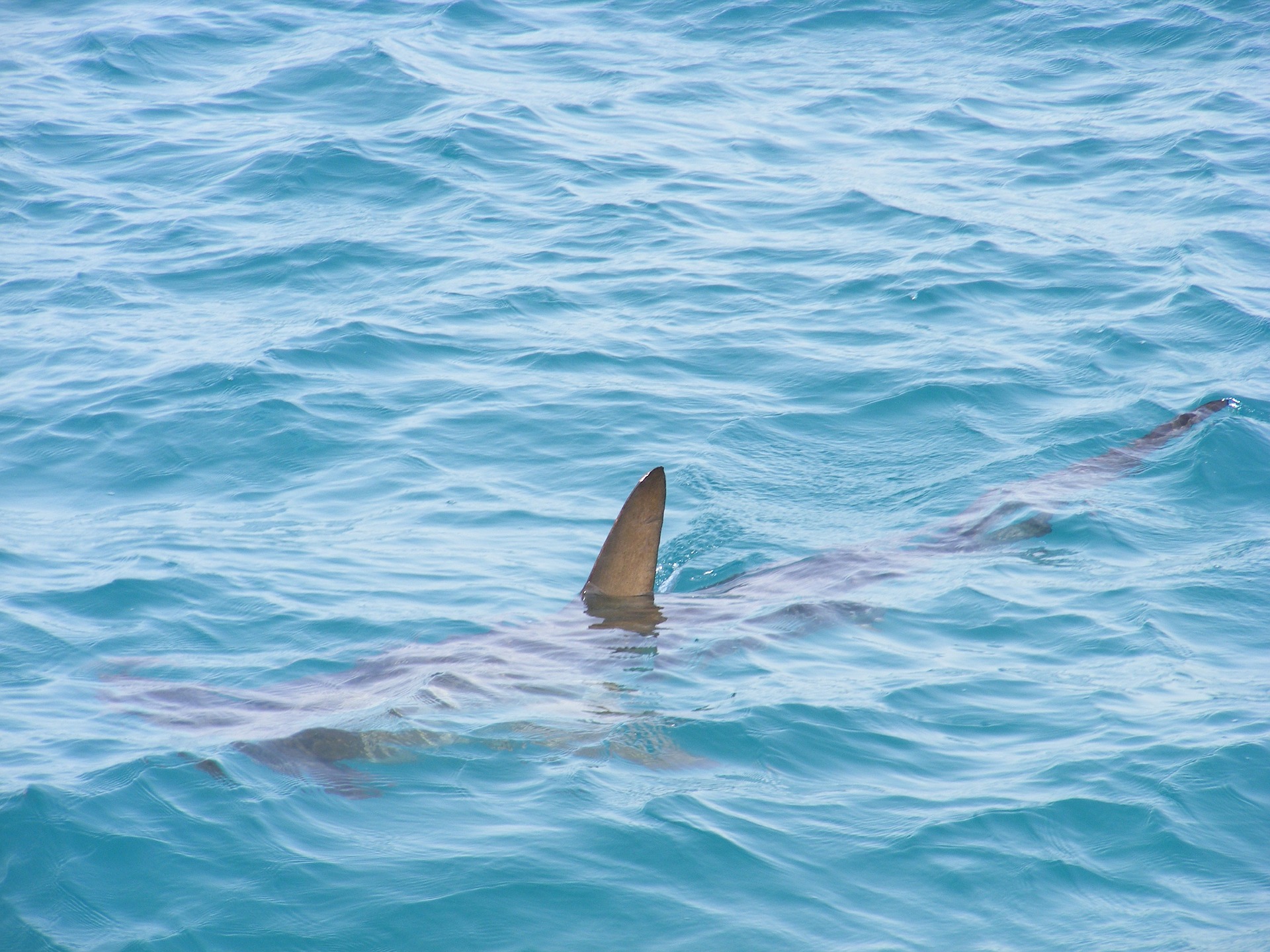}
\includegraphics[width=0.32\textwidth]{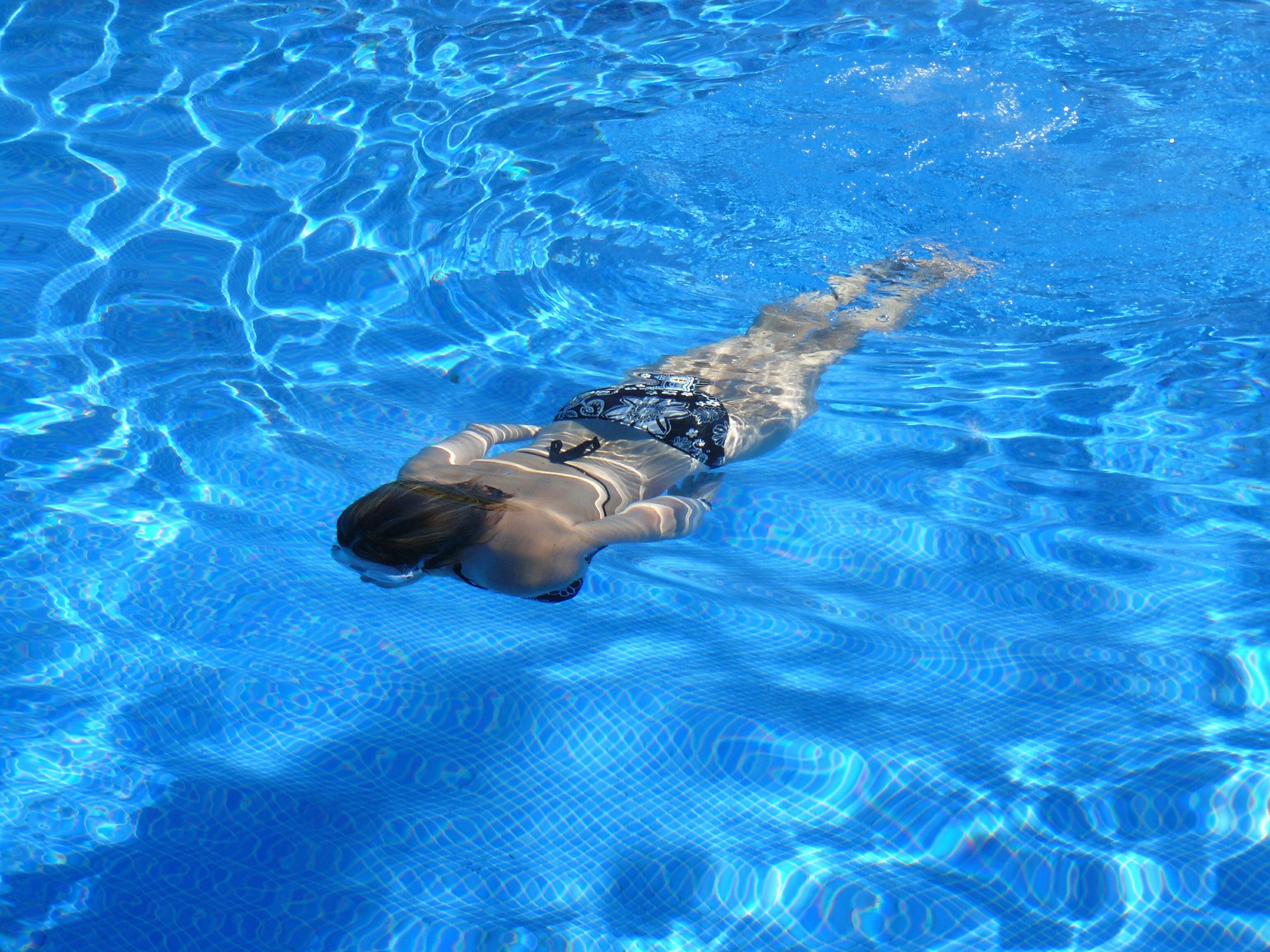}
\includegraphics[width=0.32\textwidth]{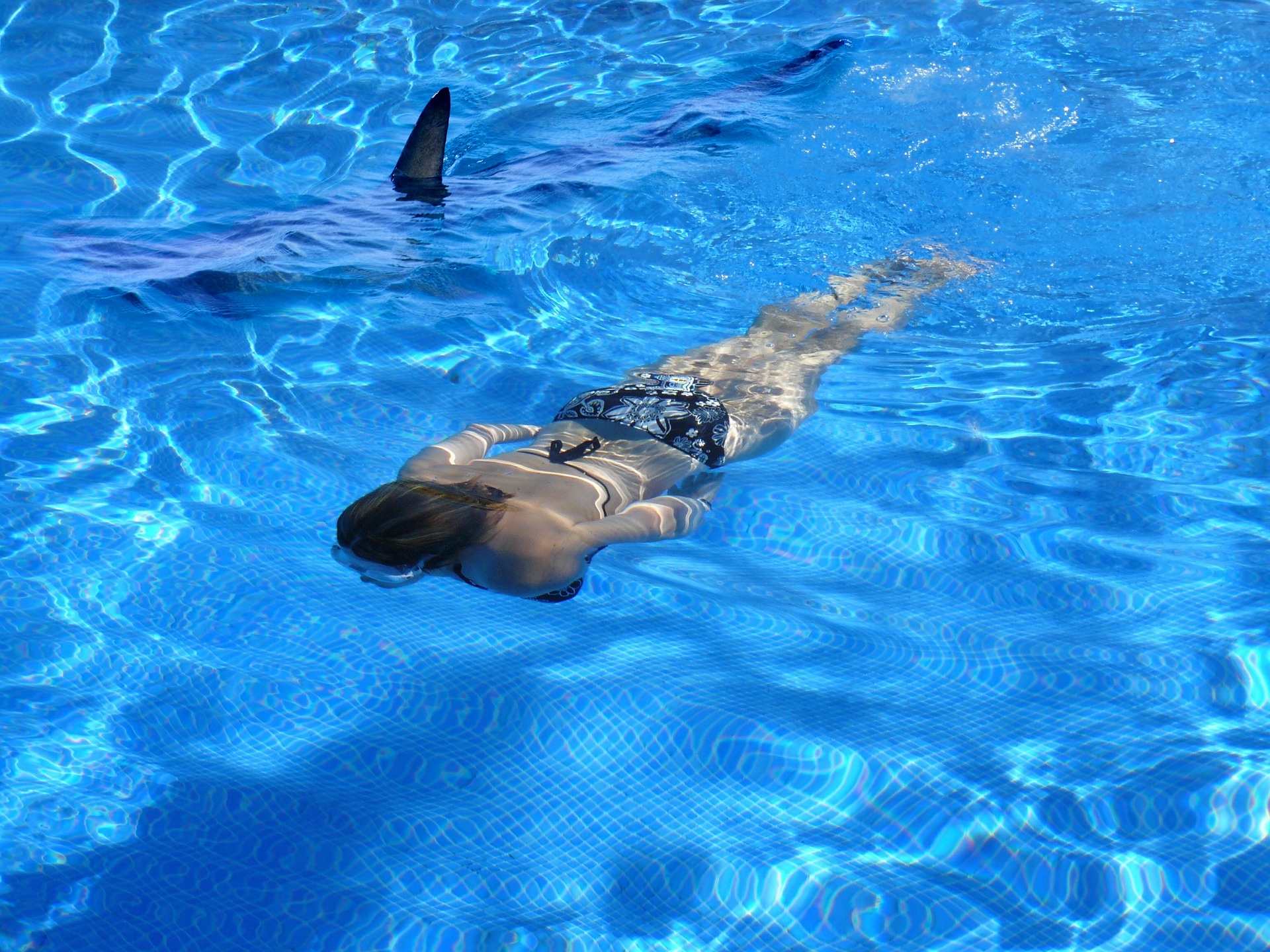}
\caption{Inserting the shark from the left image into the middle image via spectral image fusion yields the image on the right: by keeping low frequencies from the middle image, one obtains highly believable colors in the fusion result. A smooth transition between the inserted object and the background image by a fuzzy segmentation mask (alpha-matting) was used to further avoid fusion artifacts.}
     \label{fig:shark}
\end{figure}

Finally, we want to highlight that the nonlinear image fusion framework has applications beyond facial image manipulation. Similar to Poisson image editing \cite{poissonImageEditing}, one can insert objects from one image into the other by keeping low frequencies (colors and shadows) from one image and using higher frequencies (shapes and texture) from another image. Figure \ref{fig:shark} shows an example of fusing the images of a shark and a swimmer. 

\vspace{-1cm}
\subsection{Comparison to Other Techniques}
\begin{figure}[h!]
  \centering
    \subfloat[Original]{
    \includegraphics[width=0.32\textwidth]{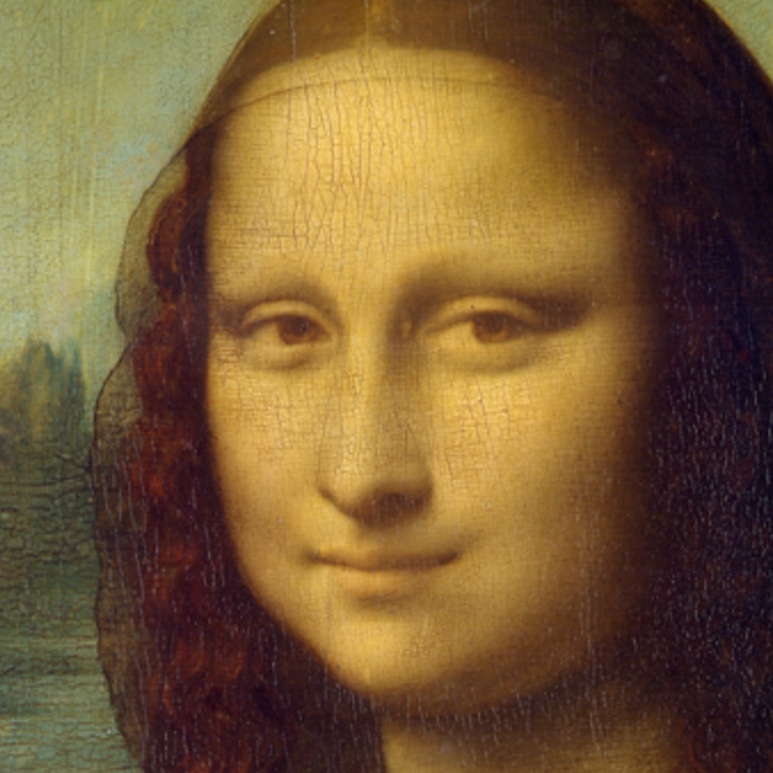}
    }
    \subfloat[Spectral]{
    \includegraphics[width=0.32\textwidth]{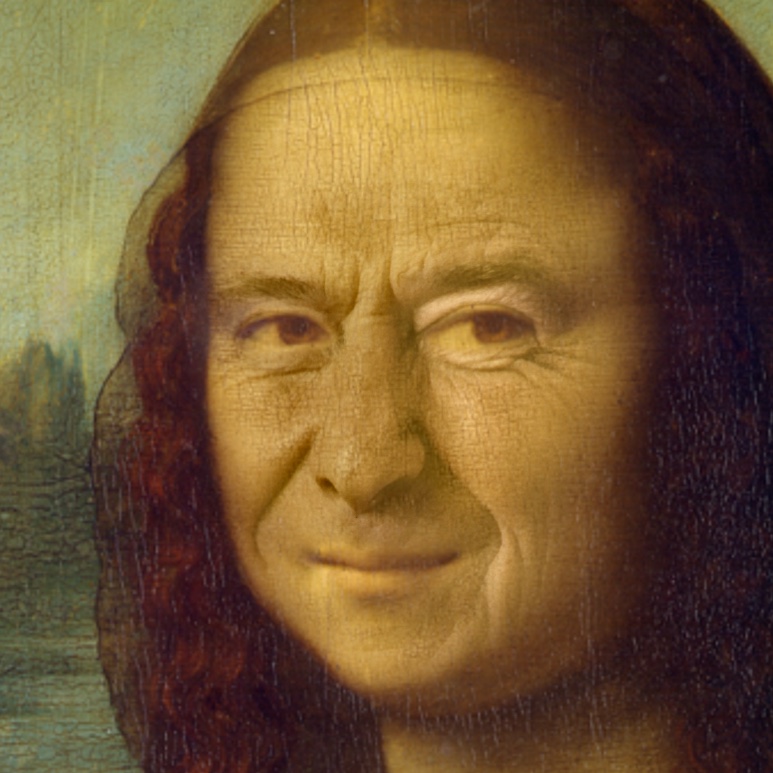}
    }
   \subfloat[Laplacian]{
    \includegraphics[width=0.32\textwidth]{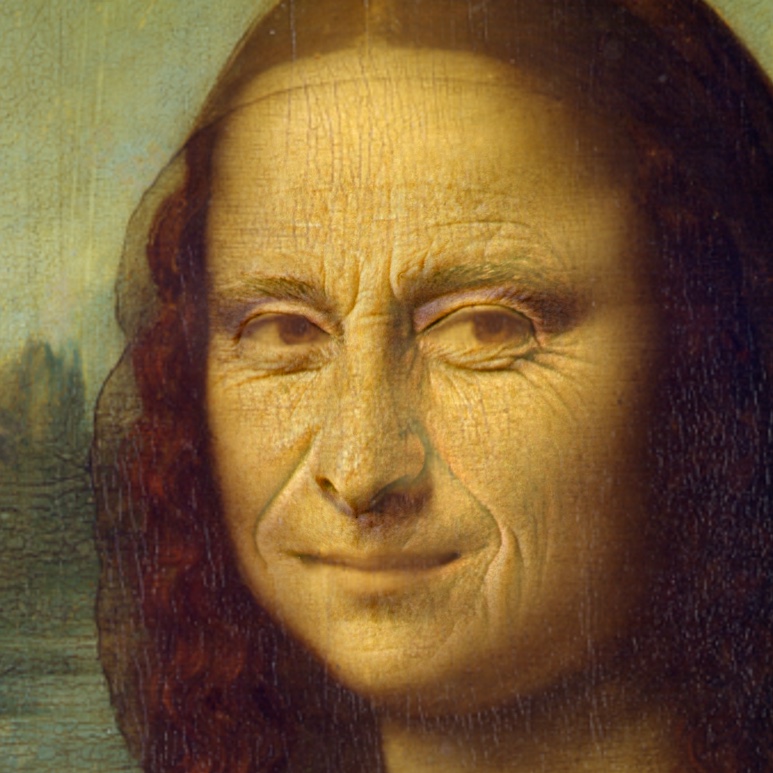}
    }\\
        \subfloat[Wavelet]{
    \includegraphics[width=0.32\textwidth]{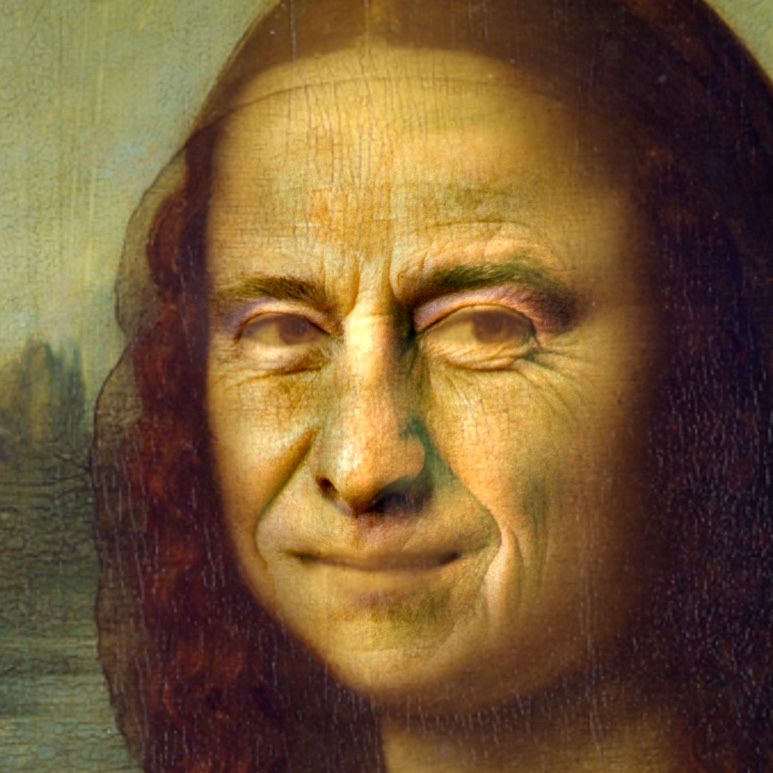}
    }
    \subfloat[Poisson]{
    \includegraphics[width=0.32\textwidth]{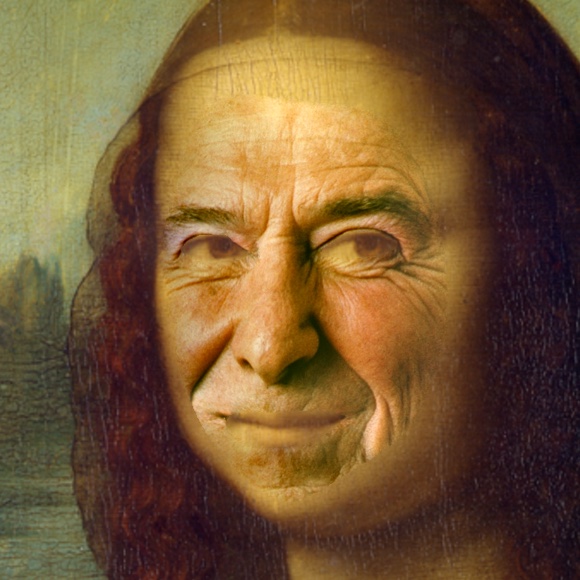}
    }
   \subfloat[Osmosis]{
    \includegraphics[width=0.32\textwidth]{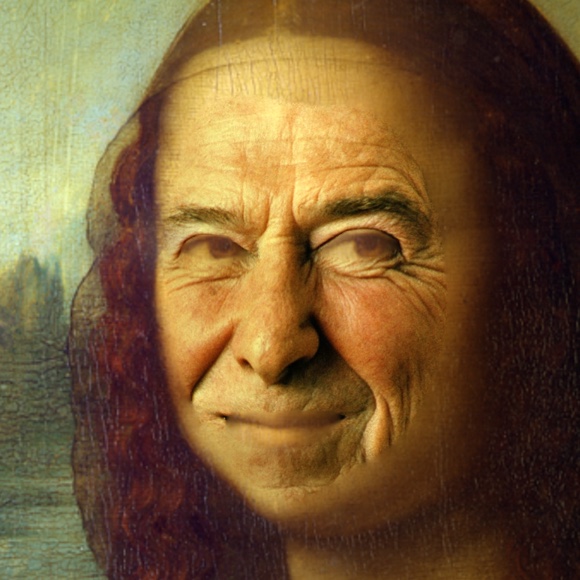}
    }
    \caption{Comparing the spectral image fusion (b) for the example of injecting details from the Reagan image into the image of Mona Lisa (a) with Laplacian pyramid fusion (c), wavelet image fusion (d), Poisson editing (e), Osmosis (e).}
    \label{fig:comparison2}
    \end{figure}

To illustrate the advantages of the image-adaptive nonlinear spectral decomposition we compare our algorithm to the classical multiscale methods of wavelet fusion, Laplacian pyramid fusion, and to the fotomontage techniques of Poisson image editing \cite{poissonImageEditing} as well as linear osmosis image editing \cite{linearOsmosis,osmosisFilters}. We compare all methods on the challenging example of fusing a photo of Reagan with the painting of Mona Lisa, see Figure \ref{fig:comparison2}. All methods use the identical registration- and segmentation-results from the automatic fusion pipeline described in Section \ref{sec:facesresults}. As we can see, Poisson and osmosis imaging transfer too many colors of the reference images and require more sophisticated methods for generating a guidance gradient field to also incorporate fine scale details of the target image such as the scratches on the painting. Wavelet image fusion generates unnatural colors and the Laplacian pyramid approach contains some halos. In particular, the texture of Reagans cheeks makes the Laplacian pyramid fusion look unnatural. By damping the filter coefficients of the nonlinear spectral decomposition, one can easily generate a result which is subtle enough to look realistic but still have clearly visible differences.

\subsection{Artistic Image Transformations}
\label{sec:materials}

Another application that demonstrates the variety of possibilities using nonlinear spectral decompositions for image manipulation is transforming an image such that the transformed image has a new look and feel. This means the image still keeps the same salient objects or features of the original image after the manipulation process, but they now seem as if they were composed in a different way.

\begin{figure}[!ht]
  \centering
  \subfloat{
    \includegraphics[width=0.23\textwidth]{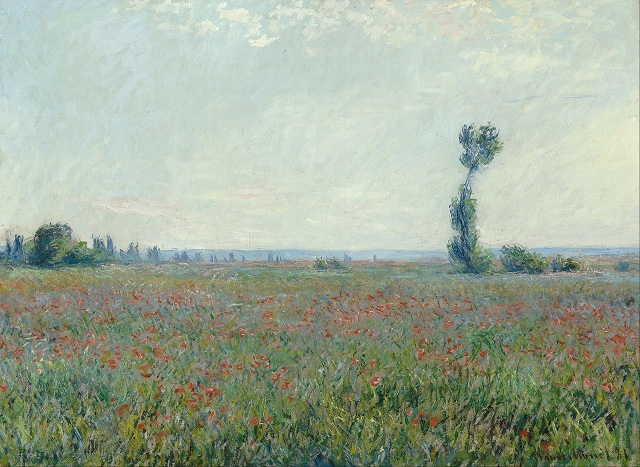}
    }
   \subfloat{
    \includegraphics[width=0.24\textwidth]{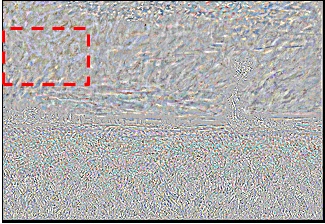}
    }
  \subfloat{
    \includegraphics[width=0.24\textwidth]{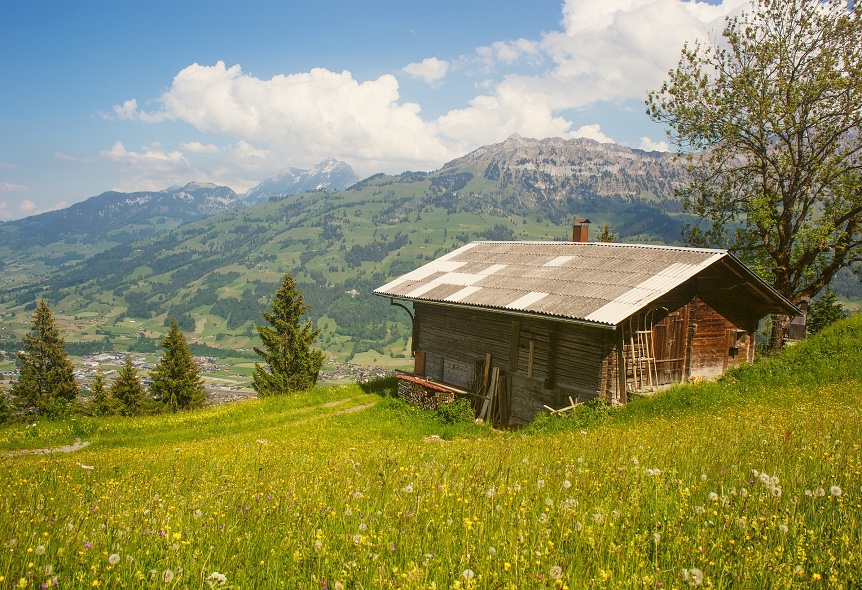}
    }
  \subfloat{
    \includegraphics[width=0.24\textwidth]{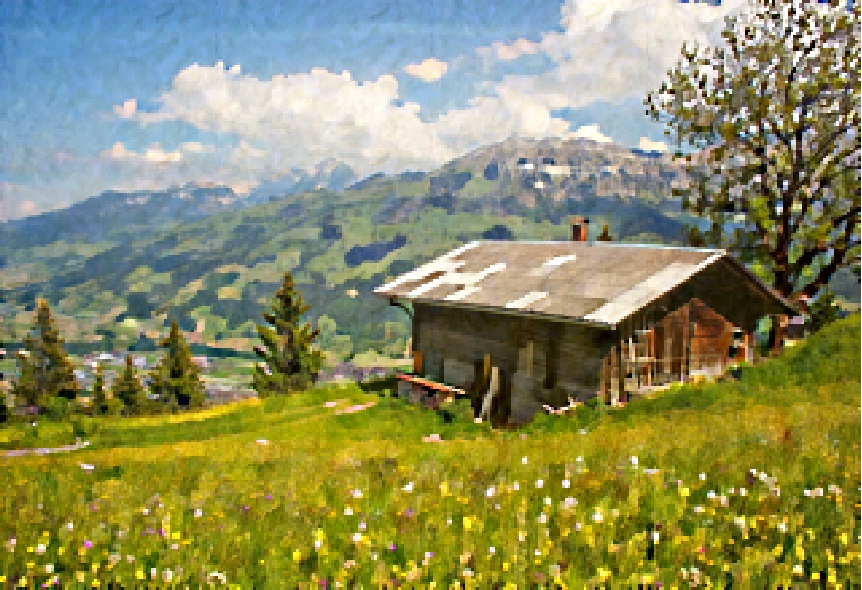}
    }
\caption{Example of transforming a photo such that it gives the impression of being an impressionist painting. Using the spectral decomposition, we extract the brush stroke features of the painting (very left) from high frequency bands at a certain area (marked in red) and embedded them into the photo image. The result is shown on the right.}
     \label{fig:cabin2paint}
\end{figure}

As a first example we consider transferring an image of a real world scene into a painting. To accomplish the latter, we extensively enhance medium frequency bands to acquire some characteristics associated with oil paintings: a small smearing effect and high contrast between different objects. To further increase the painting effect we borrow brush stroke qualities from an actual painting (Figure \ref{fig:cabin2paint} left) and combine them with the original photo. The right image in Figure \ref{fig:cabin2paint} illustrates the result of such a procedure.

\begin{figure}[!ht]
  \centering
  \subfloat{
    \includegraphics[width=0.33\textwidth]{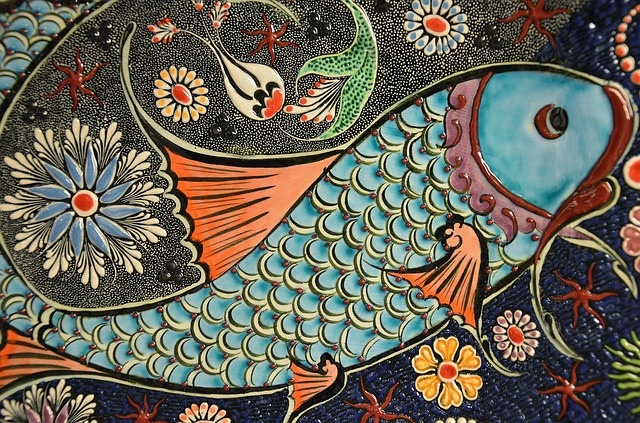}
    }
  \subfloat{
    \includegraphics[width=0.33\textwidth]{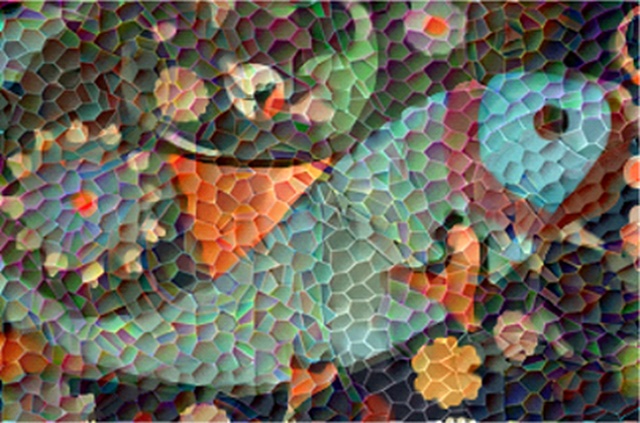}
    }
    \subfloat{
    \includegraphics[width=0.3\textwidth]{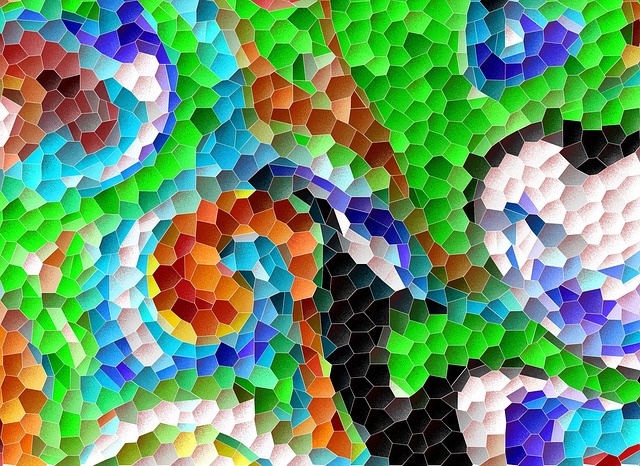}
    }
\caption{Example of transforming ceramic art such that it gives the impression of being a mosaic. Low frequencies from the ceramic art (left) are extracted and combined with the high frequencies from the mosaic image (right).}
     \label{fig:ceramic2mosiac}
\end{figure}

Figure \ref{fig:ceramic2mosiac} demonstrates a different type of manipulation enabled by nonlinear spectral decomposition. In this case we keep only very low frequencies from a fish image, and import all other frequencies from a mosaic image, leading to the impression of a fish-mosaic in the fused image.

\section{Conclusions and Future Research}
\label{sec:conclu}
In this paper we demonstrated the potential of nonlinear spectral decompositions using TV regularization for image fusion. In particular, our facial image fusion pipeline produces highly realistic fusion results transferring facial details such as wrinkles from one image to another. It provides a high flexibility, leading to results superior to methods such as Poisson image editing, osmosis, wavelet fusion or Laplacian pyramids on challenging cases like the fusion of a photo and a painting. Furthermore, it easily extends to several other image manipulation tasks, including inserting objects from one image into another as well as transforming a photo into a painting.

Note that the proposed image fusion framework is not only complementary to other image fusion techniques, but can also be combined with those, e.g. by applying them on individual bands of the spectral decomposition, which is a direction of future research we would like to look into. Further directions of future research include learning a regularization that is possibly even better suited at separating facial expressions and wrinkles from the image than the total variation.

\textbf{Data Statement:} the corresponding programming codes will be made available at \url{https://doi.org/10.17863/CAM.8305}

\section*{Acknowledgements}
MBe and CBS acknowledge support from EPSRC grant ’EP/M00483X/1’ and the Leverhulme Trust project ’Breaking the non-convexity barrier’. MBe further acknowledges support from the Leverhulme Trust early career fellowship "Learning from mistakes: a supervised feedback-loop for imaging applications" and the Newton Trust. MM acknowledges support from the German Research Foundation (DFG) as part of the research training group GRK 1564 Imaging New Modalities. RZN and GG acknowledge support by the Israel Science Foundation (grant 718/15). MBu acknowledges support by ERC via Grant EU FP 7 - ERC Consolidator Grant 615216 LifeInverse. DC acknowledges support from ERC Consolidator Grant “3D Reloaded”. CBS further acknowledges support from EPSRC centre ’EP/N014588/1’, the Cantab Capital Institute for the Mathematics of Information, and from CHiPS (Horizon 2020 RISE project grant). 

\bibliographystyle{plain}
\bibliography{refs}
\end{document}